# F4D: Factorized 4D Convolutional Neural Network for Efficient Video-level Representation Learning

Mohammad Al-Saad, Lakshmish Ramaswamy and Suchendra Bhandarkar
School of Computing
The University of Georgia
Athens, GA
{mohammad.alsaad, laksmr, suchi}@uga.edu

*Abstract*— Recent studies have shown that video-level representation learning is crucial to the capture and understanding of the long-range temporal structure for video action recognition. Most existing 3D convolutional neural network (CNN)-based methods for video-level representation learning are clip-based and focus only on short-term motion and appearances. These CNN-based methods lack the capacity to incorporate and model the long-range spatiotemporal representation of the underlying video and ignore the long-range video-level context during training. In this study, we propose a factorized 4D CNN architecture with attention (F4D) that is capable of learning more effective, finer-grained, long-term spatiotemporal video representations. We demonstrate that the proposed F4D architecture yields significant performance improvements over the conventional 2D, and 3D CNN architectures proposed in the literature. Experiment evaluation on five action recognition benchmark datasets, i.e., Something-Something-v1, Something-Something-v2, Kinetics-400, UCF101, and HMDB51 demonstrate the effectiveness of the proposed F4D network architecture for video-level action recognition.

## I. INTRODUCTION

In an era dominated by digital mediums, the increasing number of large-scale videos has transformed the way information is conveyed and consumed. From autonomous vehicles and intelligent surveillance systems to online streaming services and social media platforms, videos have emerged as a pervasive and rich source of data that captures the essence of human experiences and surrounding environment. Nevertheless, the complexity and sheer volume of these huge videos present the demand for effective video understanding. The initial step of the video understanding is action recognition which aims to interpret and understand human actions, gestures, and movements.

Many 2D and 3D Convolutional Neural Network (CNN) architectures have been proposed for the problem of video-based human action recognition. A straightforward CNN-based approach to this problem uses the entire video as an input to the CNN followed by a fully convolutional inference [1]. However, the data volume in videos are huge which could result in a very high memory footprint and processing power as trying to run a fully convolutional inference is well above the capabilities of modern GPUs [2].

To substantially reduce the memory footprint and computational cost, most existing deep learning (DL) models for video representation learning incorporate clip-level feature learning This allows these DL models to apply deep networks over video clips of fixed temporal length focusing on short-term object appearances and motion, thus, learning from video clips instead of the entire video. The clip-based learning methods sample short video clips comprising of 10-32 frames per clip, and compute the prediction scores for each clip independently [3]. Finally, the individual results from all the clips are pulled together to generate a final video-level prediction.

In general, clip-based models often ignore long-range spatiotemporal dependencies and the global video-level structure during training. The temporal dependency problem in vision-based human action recognition refers to the challenge of correctly capturing and modeling the dynamic and sequential nature of human actions over time. It identifies that actions are not separated events but unfold as a sequence of distinctive motion patterns, each pattern contributes to the overall understanding of the action being performed. Temporal dependency holds the notion that the duration, timing, and order of these motion patterns are critical for interpreting and recognizing actions correctly. Capturing the temporal aspect is essential for distinguishing between actions that may share similar visual appearance but vary in their execution timing or sequence. In many cases, partial observation of the underlying video makes it very difficult to recognize an action correctly. Additionally, relying on the average of the prediction scores from individual clips is considered to result in a sub-optimal inference.

To learn from an entire video efficiently, the Temporal Segment Network (TSN) architecture has been proposed [4]. The TSN represents the contents of the entire video by operating on a sequence of multiple short clips (snippets) sampled from the entire video. In the final TSN stage, a segmental consensus function is used to aggregate the predictions from the sampled snippets, thereby enabling the TSN to model long-range temporal structures. However, the fact that inter-clip interactions and video-level fusion are performed in the final TSN stage limits the ability of the TSN to capture fine temporal structures. To overcome this limitation, the V4D CNN model [5] incorporated the 4D CNN architecture. The 4D convolution operation has the capacity to model long-range dependencies and capture inter-clip interactions for efficient video-level representation learning. To capture finer temporal structures, the V4D CNN residual blocks are placed at earlier stages in the network. Nevertheless, the 4D convolution operation in the V4D CNN model is complex and introduces many more parameters thereby making the model vulnerable to overfitting. Furthermore, the V4D CNN architecture does not incorporate an attention mechanism to focus on the regions of interest (ROIs) that evolve over time.

Inspired by the above observations of the state of the art in video-level representation learning, we propose an effective yet simple framework for video level representation learning termed as the Factorized 4D (F4D) architecture, to model both short-range motion and long-range temporal dependency within a large-scale video sequence. This paper has two main objectives; the first objective is to enhance accuracy and to decrease the complexity of the 4D convolution operation introduced in the V4D CNN framework. We start by factorization of the 4D convolution operation which renders the proposed F4D CNN model capable of representing more complex functions by capturing more complex inter-clip interactions and finer temporal structures. Furthermore, the proposed factorization improves the optimization procedure during both training and testing, yielding lower training and testing errors. The second objective is to implement an attention mechanism that focuses on an ROI within the video and enhances the power of the resulting representation. We design two attention mechanisms, namely the temporal attention (TA) module and spatio-temporal attention (STA) module. These modules will focus on the different inter-clip motion patterns that evolve over time and on the spatio-temporal discriminative features by focusing on the ROIs that evolve over time. We insert the proposed factorized 4D CNN followed by the attention modules to form a block named F4D residual block. The F4D residual blocks can be easily inserted into standard ResNet [6] architecture to form the F4D architecture. The main contributions of our work can be summarized as follows:

- We propose a Factorized 4D CNN that can capture more complex long-range temporal dependency and inter-clip interactions with lowered training and testing errors compared to the 4D CNN.

- We propose a temporal attention module (TA) and a spatio-temporal attention module (STA) that guide the network to focus on ROIs within the video and improves the resulting representation with negligible computation cost.

- An effective yet simple network referred as F4D architecture is proposed with our F4D residual blocks that consist of the proposed F4D CNN followed by the proposed attention modules, which can be easily integrated into standard ResNet architecture.

- Extensive experiments demonstrate the effectiveness of the proposed F4D architecture on five action recognition benchmark datasets including Something-Something-v1 and v2 [7], Kinetics-400 [8], UCF101 [9] and HMDB51 [10].

## II. RELATED WORKS

**Two-Stream 2D CNN.** The two-stream CNN architecture represents a very practical approach to video-level representation learning. The earliest two-stream CNN architecture was introduced in [11] where one CNN learns from a stream of RGB frames and the other CNN from a stream comprising of stacks of 10 computed optical flow frames. In the later stages, the results of both streams are averaged to yield the final prediction.

Although the two-stream CNN architecture has been shown to yield impressive results, the extraction of spatial and temporal features is performed independently, and it is easy to ignore their intrinsic connection, which can influence the final prediction. Another limitation of two-stream networks is the excessive demands of optical flow computation where parallel optimization is difficult to implement. Some related works have explored the idea of enhancing the optical flow computations [12,13,14,15] in this regard.

**3D CNN.** Since 3D CNNs incorporate spatio-temporal filters, they represent a natural approach to video modelling. The biggest advantage of 3D CNNs is their ability to create hierarchical representations of spatio-temporal data. 3D CNNs have been explored in several works cited in the literature. Ji et al. [16] pioneered the use of the 3D CNN for human action recognition by applying 3D convolution operation in both the spatial and temporal domains.

Tran et al. [17] propose the C3D model and show its effectiveness when trained on large-scale video datasets. They conducted a systematic study to show that 3D CNN is better than 2D CNN in learning appearance and motion information. Moreover, they show that using 3×3×3 convolution kernels for all layers works best amongst the explored architectures. The work in [18] improves upon the C3D model by employing neural architecture search across multiple dimensions and 3D residual networks that allow for use of deeper networks that can be trained on large-scale video datasets.

The two-stream 3D CNN architecture has been explored by Carreira et al. [19] with the goal of successfully incorporating 2D image classification models into a 3D CNN by inflating all the filters and pooling kernels by adding an extra temporal dimension. The authors use a pre-trained Inception framework as the architectural backbone with one stream trained on RGB inputs and another stream trained on optical flow. Recent work in [20] improves the 3D residual architecture by decoupling the 3D convolutional kernel and also presents the design of a 3D attention mechanism to decrease the model's sensitivity to changes in the background environment.

There are several disadvantages associated with the 3D CNN architecture. First, the number of 3D CNN model parameters increases more rapidly compared to the 2D CNN. Second, the 3D CNN is hard to train and the resulting training information hard to transfer, and its inference process very slow compared to other approaches. Third, in some cases, the 3D convolution operation cannot distinguish between the human action features and the background features making the model vulnerable to environmental factors.

**Mapping from 2D to 3D CNN.** Several research papers have explored techniques to transfer the benefits of pre-trained 2D CNNs to 3D CNN architectures. In [21], the authors consider the 2D Resnet and replace all its 2D convolutional filters with 3D convolutional kernels to arrive at the ResNet3D architecture. They assume that a combination of large-scale datasets and deep 3D CNNs are capable of replicating the success of 2D CNNs on the ImageNet dataset. Inspired by ResNeXt architecture [22], Chen et al. [23] propose a multi-fiber architecture that divides a complex neural network into an ensemble of lightweight networks thereby reducing the Identify the computational cost and simultaneously coordinating the information flow. Motivated by the SENet [24], the STCNet architecture [25] incorporates channel-wise information within a 3D block to capture the correlation information between the temporal and spatial channels throughout the network.

**Unifying 2D and 3D CNN.** 3D CNNs have witnessed great success in recognizing human action in videos. However, the high complexity of training the 3D convolution kernels and the need for large quantities of training videos limits their applicability. To reduce the complexity of 3D CNN training, the P3D [26] and R(2+1)D [3] architectures explore the idea of 3D factorization wherein a 3D kernel is factorized into two separate operations, a 2D spatial convolution and a 1D temporal convolution. Trajectory convolution [27] is based on a similar concept but utilizes deformable convolution for the temporal component to better deal with motion. A different approach of simplifying 3D CNNs is to integrate 2D and 3D convolutions within a single network. MiCTNet [28] integrates 2D and 3D CNNs to generate richer, deeper, and more informative feature maps by decreasing the complexity of training in each round of spatial-temporal fusion. ARTNet [29] establishes a relation and appearance network by using a novel building block comprising of a spatial branch using 2D CNNs and a relation branch using 3D CNNs. S3D [30] and ECO [31] combine the advantages of the aforementioned models by adopting a top-heavy network to achieve online video understanding.

**Long-term Video Modelling Frameworks.** In their seminal work, Wang et al. [4], propose a simple, flexible, and general framework for learning action models in videos. *Temporal segment networks* (TSNs) are designed by performing sparse sampling of a long video to extract short snippets followed by a segmental consensus function to aggregate information from the sampled snippets. This allows the TSN to model long-range temporal structures within the entire video. The *Temporal Relational Reasoning Network* (TRN) [32] enables temporal relational reasoning over videos by describing the temporal relations between observations in videos. While the TRN is shown to be capable of discovery and learning of potential temporal relations at multiple time scales within a video, it lacks the capacity to capture finer temporal structure. For efficient video understanding, Liu et al. [33]

introduce a *Temporal Shift Module* (TSM) that extends the shift operation to design a temporal module to capture temporal relations. The STM architecture [34] incorporates two channel-wise modules, one to represent motion features and the other to encode spatio-temporal features. Inspired by the approach in [24], the TEA architecture [35] improves the motion pattern representation by using the motion features to calibrate the spatio-temporal features.

**4D CNN**. The V4D CNN architecture proposed by Zhang et al. [5] tackles the analysis of RGB videos by incorporating a video-level sampling strategy to cover the holistic duration of a given video. A novel 4D residual block is proposed which allows the casting of 3D CNNs into 4D CNNs for learning long-range interactions of the 3D features, resulting in a "time of time" video-level representation. The proposed V4D architecture has achieved excellent results compared to its 3D counterparts.

### III. F4D ARCHITECTURE

#### A. Segment Based Sampling

To model the long range spatio-temporal dependency, we use segment-based sampling described in [4]. Formally, given a whole video $V$, we divide it into $U$ sections of equal durations and select a snippet, termed as an *action unit*, that is randomly sampled from each section to represent a short-term action pattern within that section. The holistic action in the video is represented by a sequence of action units $\{A_1, A_2, ..., A_U\}$, where $A_i \in \mathbb{R}^{C \times T \times H \times W}$ is the action unit obtained from the $i^{th}$ section, $C$ is the number of channels, $T, H, W$ are the temporal length, height, and width . During the training phase, each action unit $A_i$ is randomly selected from each of the $U$ sections. During testing, the center of each $A_i$ is located exactly at the center of the corresponding section.

#### B. Overview of 4D CNN

In recent years, the 3D CNN has been shown to be a powerful approach for modelling short-term spatio-temporal features in video. However, the receptive fields of 3D kernels are usually deficient owing to the compact sizes of kernels, and hence pooling operations are applied to enlarge the receptive fields. In contrast, 4D convolution operations have been implemented to simultaneously model short-term and long-term spatio-temporal representations since they have the capacity to model long-range dependencies and capture inter-clip interactions for efficient video-level representation learning.

The input to a 4D convolution can be denoted as a tensor V of size $(C, U, T, H, W)$, where $U$ is the number of action units (the 4th dimension). The batch dimension has been excluded for simplicity. Formally, a 4D convolution operation can be viewed as follows:

$$o_j^{uthw} = b_j + \sum_c^{C_{in}} \sum_{s=0}^{S-1} \sum_{p=0}^{P-1} \sum_{q=0}^{Q-1} \sum_{r=0}^{R-1} W_{jc}^{spqr} v_c^{(u+s)(t+p)(h+q)(w+r)} \quad (1)$$

where $o_j^{uthw}$ is a pixel at position $(u, t, h, w)$ of the $j^{th}$ channel in the output following the annotation in [36], $b_j$ is a bias term, $c$ is one of the $C_{in}$ input channels of the feature maps, $S \times P \times Q \times R$ is the shape of 4D convolutional kernel, $W_{jc}^{spqr}$ is the weight at the position $(s, p, q, r)$ of the kernel, corresponding to the $c^{th}$ channel of the input feature maps and $j^{th}$ channel of the output feature maps. Since deep learning libraries do not provide an implementation for 4D convolutions, eqn. (1) can be modified to generate eqn. (2) which allows the implementation of 4D convolutions using 3D convolutions. Eqn. (2) can be formulated as follows:

$$o_j^{uthw} = b_j + \sum_{s=0}^{S-1}(\sum_{c}^{C_{in}}\sum_{p=0}^{P-1}\sum_{q=0}^{Q-1}\sum_{r=0}^{R-1} W_{jc}^{spqr} v_c^{(u+s)(t+p)(h+q)(w+r)}) \quad (2)$$

where the expression in the parentheses can be implemented by 3D convolutions. Within the 4D space, the 4D convolution kernel has the ability to model both the short-term 3D features of each action unit and the long-term temporal evolution of several action units at the same time. Thus, the 4D convolutions have the power to learn more complicated interactions of a long-range 3D spatio-temporal representation.

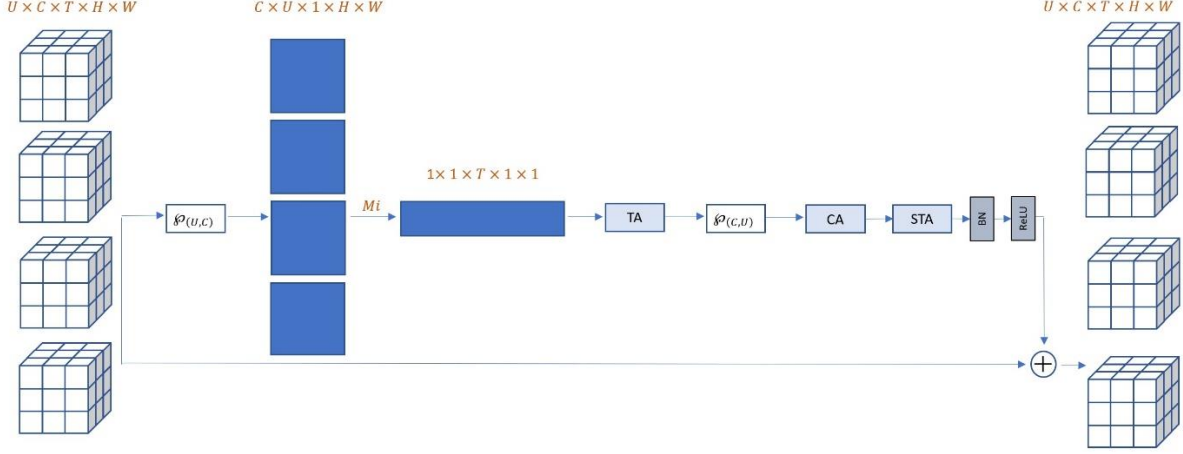

Fig. 1. F4D Residual Block.

### C. F4D: Factorization of 4D CNN

In this section, we design a network block termed as F4D to improve upon the 4D convolution discussed in the previous section. We follow the work in [3] to approximate the 4D convolution by a 3D convolution followed by a 1D convolution, thereby decomposing the spatial modeling and the temporal modeling for action units into two separate steps. The (3+1)D block replaces the $N_i$ 4D convolutional filters of size $N_{i-1} \times u \times t \times h \times w$, with $M_i$ 3D convolutional filters of size $N_{i-1} \times u \times 1 \times h \times w$ and $N_i$ temporal convolution filters of size $M_i \times 1 \times t \times 1 \times 1$. The hyperparameter $M_i$ decides the dimensionality of the intermediate subspace where the signal is projected between the spatial convolution and the temporal convolution. In order to have a (3+1)D block with the number of parameters approximately equal to the number of parameters in the implementation of a full 4D convolution layer, we set $M_i = \left\lfloor \frac{u\,t\,h\,w\,N_{i-1}\,N_i}{u\,h\,w\,N_{i-1} + tN_i} \right\rfloor$.

The (3+1)D decomposition provides advantages over the full 4D convolution. First, although the number of parameters is approximately the same, the number of nonlinearities in the F4D network will increase due to the additional ReLU between the 3D and the 1D convolution in each block. Adding more nonlinearities results in increased complexity of functions that can be represented. This has been noted in VGG [37] and R(2+1)D networks which approximate the effect of a big filter by applying several smaller filters with additional nonlinearities introduced between them. Second, forcing the 4D convolution into separate spatial and temporal modules can render the optimization easier, resulting in lower training error compared to the 4D convolution of the same size and capacity. Hence, for the same number of layers and parameters, the (3+1)D block will have lower training error and lower testing error compared to the V4D network. Despite the fact that (3+1)D is a simpler architecture, experimental results show that it significantly outperforms the V4D network.

### D. F4D block Integration

This section discusses the ability of integrating the F4D blocks into existing state-of-the-art 3D CNN frameworks for action recognition. As in [5], we design a factorized 4D convolution in the residual structure [6], which shows the efficacy of combining the short-term 3D features and the long-term spatiotemporal representations for video action recognition. We start by defining a permutation function $\wp(d_i, d_j): A^{d_1 \times \ldots \times d_i \times \ldots \times d_j \times \ldots d_n} \mapsto A^{d_1 \times \ldots \times d_j \times \ldots \times d_i \times \ldots d_n}$, which permutes the dimensions $d_i$ and $d_j$ of a tensor $A \in \mathbb{R}^{d_1 \times \ldots \times d_n}$. Formally, the residual factorized 4D convolution block can be formulated as:

$$\mathcal{Y}_{3D} = \mathcal{X}_{3D} + \wp_{(U,C)}(\mathcal{F}_{3D} + \mathcal{F}_{1D}(\wp_{(C,U)}(\mathcal{X}_{3D}); \mathcal{W}_{3D} + \mathcal{W}_{1D})) \quad (3)$$

where $\mathcal{F}_{3D} + \mathcal{F}_{1D}(\mathcal{X}; \mathcal{W}_{3D} + \mathcal{W}_{1D})$ is the factorized 4D convolution operation, and $\mathcal{Y}_{3D}, \mathcal{X}_{3D} \in \mathbb{R}^{U \times C \times T \times H \times W}$. In order to process $\mathcal{X}_{3D}, \mathcal{Y}_{3D}$ using standard 3D CNNs, $U$ is merged into the batch dimension whereas in order to process $\mathcal{X}_{3D}$ using the factorized 4D convolution, we utilize the permutation function $\wp$ to permute the dimensions of

$\mathcal{X}_{3D}$ from $U \times C \times T \times H \times W$ to $C \times U \times T \times H \times W$. Thus, the output of the factorized 4D convolution can be permuted back to the 3D form so that the output dimensions are consistent. The factorized 4D convolution is followed by a batch normalization layer [38], ReLU activation and a dropout layer. In theory, any 3D CNN architecture can be recast as a factorized 4D convolution using the proposed residual block.

*E. Attention in F4D Blocks*

Inspired by CBAM network [39], we implement two attention modules and embed it within the F4D block to learn better and more refined long-term spatiotemporal representations with negligible computation overhead. The proposed attention has three major components: temporal attention map over all action units, the channel attention map, and the spatio-temporal attention map. We arrange the attention modules by placing the temporal attention map in the 4D space, and both the channel attention map and the spatio-temporal attention map after permuting back to the 3D dimension.

**Temporal Attention (TA) Map.** In order to concentrate on the long-term temporal evolution of all action units, we design a temporal attention map that focuses on the different inter-clip motion patterns that evolve over time. Given an intermediate feature map $F \in \mathbb{R}^{C \times U \times T \times H \times W}$ as input, we infer a temporal attention map $M_T \in \mathbb{R}^{1 \times U \times T \times 1 \times 1}$ by utilizing both average pooling and max pooling along the channel and spatial dimensions to obtain two feature descriptors $F_{avg}^T$ and $F_{max}^T$. Although CBAM network adopts a filter size of 7×7 which is considered a design choice that has low computation cost in 2D image-related tasks, using a convolutional operation with such a large filter size in 3D or 4D space incurs a significant computational cost in our model. To obtain substantial computational cost savings, we use the dilated convolution. We adopt a two-path 1D dilated temporal convolution [40]. The first path has a temporal dilated convolution with a dilation factor = 2 (skipping 1 pixel). The second path has a temporal dilated convolution with a dilation factor = 3 (skipping 2 pixels). The two paths model the multiscale global temporal interdependency between all action units. The temporal attention map is computed as follows:

$$M_T(F) = \sigma(Conv1D\,([AvgPool(F) + (MaxPool(F)])) \quad (4)$$

$$= \sigma\left(Conv1D\left(\left[\left(F_{avg}^T + F_{max}^T\right]\right)\right)\right) \quad (5)$$

Where $\sigma$ denotes the sigmoid function, and $Conv1D$ denotes the multipath dilated temporal convolution layer. The refined feature map after the temporal attention module is computed as:

$$F_{TA} = M_T \otimes F + F \quad (6)$$

Where $\otimes$ denotes the element-wise multiplication, $+$ denotes the inner residual connection and $F_{TA}$ the refined feature map. In the original implementation of CBAM, feature refinement is attained by multiplying the attention maps with the input feature map. However, it does not take into consideration the preservation of the original feature map. We use inner residual connections in all attention modules to preserve the original

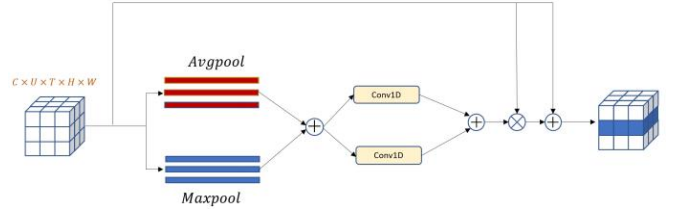

Fig. 2. Temporal Attention Module

information. This helps to avoid any unrelated features or background noise in the current layers.

**Channel Attention (CA) Map.** As in the CBAM network, the channel attention map is produced by exploiting the inter-channel relationship of features. Given an intermediate feature map $F' \in \mathbb{R}^{U \times C \times T \times H \times W}$, we compute the channel attention map by using both, the max-pooled features and average pooled features at the same time generating two different descriptors. Subsequently, both descriptors are fed to a multi-layer perceptron with one hidden layer with an activation size of $\mathbb{R}^{C/r \times 1 \times 1 \times 1 \times 1}$, where $r$ is the reduction ratio (we set $r = 16$). The output feature vectors are then combined using element-wise summation. The entire process can be summarized as follows:

$$M_C(F') = \sigma(MLP(AvgPool(F')) + MLP(MaxPool(F'))) \quad (7)$$

$$= \sigma(W_1((W_0(F_{max}'^C)) + W_1((W_0(F_{max}'^C))) \quad (8)$$

Where $W_0 \in \mathbb{R}^{C/r \times c}$ and $W_1 \in \mathbb{R}^{C \times C/r}$. In this case, the weights, $W_0$ and $W_1$ are shared by both inputs and the ReLU activation function is followed by weighting by $W_0$. The channel attention map can be summarized as follows:

$$F_C = M_C \otimes F' + F' \quad (9)$$

During multiplication, the channel attention values are copied along the spatial dimension and the temporal dimension.

**Spatio-temporal Attention (STA) Map.** This module is designed to focus on the spatio-temporal discriminative features by concentrating on the ROIs that evolve over time. The spatio-temporal attention map is generated by exploiting the inter-spatial relationship of features. Given an intermediate feature map $F' \in \mathbb{R}^{U \times C \times T \times H \times W}$, we compute the spatio-temporal attention map by first applying both, the max-pooled operations $F_{max}'^{ST} \in \mathbb{R}^{1 \times 1 \times T \times H \times W}$ and the average pooled operations $F_{avg}'^{ST} \in \mathbb{R}^{1 \times 1 \times T \times H \times W}$ along the channel axis and concatenate them to generate a refined and efficient feature descriptor $m_{ST}$. Subsequently, we forward $m_{ST}$ to a two-path 2D dilated convolution layer (with skipping 1-pixel and skipping 2-pixels) and two-path 1D dilated temporal convolution layer (with skipping 1-pixel and skipping 2-pixels). These two layers are designed to explore multiscale spatial relationships and local temporal interdependencies respectively.

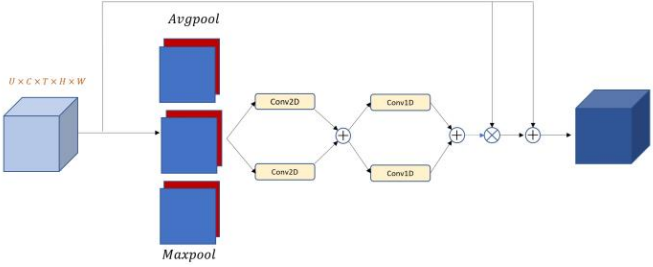

Fig. 3. Spatio-Temporal Attention Module

In summary, the spatio-temporal attention is computed as:

$$m_{ST} = Concatenate[F'^{ST}_{avg}, F'^{ST}_{max}] \qquad (10)$$

$$M_{ST} = \sigma(Conv1D(ReLU(Conv2D(m_{ST})))  \qquad (11)$$

where $Conv2D$ represents the two path 2D convolution layer. The refined feature map is computed as:

$$F_{ST} = M_{ST} \otimes F_C + F_C \qquad (12)$$

where $F_{ST}$ is the refined feature map.

## IV. EXPERIMENTS

### A. Datasets

Five benchmark datasets have been used for experimental evaluation of the proposed F4D convolution block: Something-Something-v1, Something-Something-v2 [7], Kinetics-400 [8], UCF101 [9], HMDB51 [10]. Something-Something-v1 is a dataset that contains labeled video clips of humans performing predefined actions. It consists of 108,499 videos, with 86,017 in the training set, 11,522 in the validation set and 10,960 in the testing set comprising of 174 action classes. Something-Something-v2 is an extension of the first version with a collection of 220,847 videos incorporating several enhancements such as higher video resolution, and reduced label noise. The Kinetics 400 dataset covers 400 action classes with ≈400 video clips for each action. The video clips are obtained from different YouTube videos with each video clip lasting ≈10 seconds. The actions are human focused, and the action classes include a wide range of human-human and human-object interactions. The UCF101 dataset consists of 13320 video clips with 101 action classes. This dataset includes several variations arising from multi-viewpoints, camera motion, object appearance, cluttered background, and illumination conditions. The HMDB51 dataset has 51 action classes distributed across 6849 video clips collected from different sources and public databases such as YouTube, Google and the Prelinger archive.

### B. Implementation Details

We perform our initial evaluation on Something-Something datasets, using the training split for training and the validation split for testing. To learn the network parameters, we use the mini batch stochastic gradient descent (SGD) as the optimization algorithm. The batch size is set to 128 and the momentum to 0.9. Initially, the learning rate is set to 0.01, and drops by a factor of 10 at epochs 20, 40, and 60. Model training is concluded at 80 epochs. Batch normalization is applied to all convolutional layers. We follow each F4D convolutional block with batch normalization, ReLU activation and a dropout layer. To speed up training, we utilize the data parallelism strategy implemented using the torch.nn.DataParallel module in Pytorch to split the mini-batch of samples into multiple smaller mini-batches and perform the computation over four Tesla P100-PCIE-16GB GPUs. Data augmentation plays an important role in enhancing the performance of deep learning architectures. During training, we use random left-right flipping, location jittering, scale jittering and corner cropping.

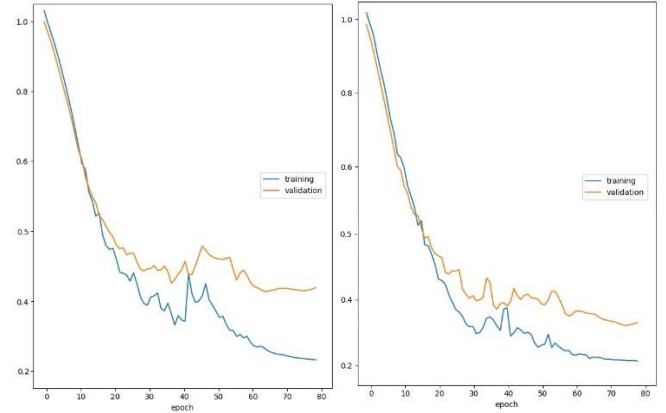

Fig. 4. Training and Testing errors for V4D (left) and F4D (right)

### C. Results on Motion-Focused Datasets

In this section, we evaluate our proposed approach with the state-of-the-art approaches on motion-focused datasets including Something-Something-v1 and Something-Something-v2. Both datasets focus on modelling motion and temporal information where the motion of actions is more complicated compared to that in the Kinetics-400 dataset albeit with a clearer background. Videos in both datasets contain one continuous action with clear start and end points along the temporal dimension. To prepare the videos for training, we use the segment-based sampling technique explained in Section 3.1. We segment the holistic duration of a video into $U$ sections of equal durations in their temporal order and for each section, we randomly select a snippet composed of 32 frames. To form an action unit, we take each snippet and use the sampling strategy mentioned in [2] to sample 8 frames with a fixed stride of 4. We also experiment with the number of frames in the snippet set to 16 with the frame size fixed at 256×256 pixels. After applying the data augmentation techniques mentioned in the previous section, we resize the cropped region to 224×224 pixels. We fix $U$=4 in all of experiments. For fair comparison, we use the ResNet50 CNN as the backbone for proposed F4D network.

For inference, we follow the approach in [2, 41] using fully spatial convolutional testing. From the entire duration of a video, we sample 10 action units ($U$=10) of equal duration, scale up the smaller spatial image dimension to 256 pixels and take 3 crops of 256×256 pixels to spatially cover the entire frame for each action unit, and then resize the crops to 224×224 pixels. Finally, the final prediction is produced via global average pooling over the sequence of all action units.

Fig.4 highlights the training error and testing error for V4D CNN and F4D architecture. It is illustrated that for the same network backbone (ResNet 50) and approximately the same number of parameters, the F4D architecture achieves lower training error and lower testing error. This shows that the factorization of the 4D CNN renders the optimization easier and achieves better resulting representation.

Fig. 5 and Fig. 6 show the results of our approach compared to the state-of-the-art approaches on the Something-Something datasets. Compared with the baseline approach that uses a TSN with 8 frames, the proposed F4D approach with 8 frames achieves a 35.2% improvement with top-1 accuracy of 54.9 with 8 frames on the Something-Something-v1 dataset when pretrained on ImageNet [42]. When the proposed F4D model is pretrained on ImageNet and Kinetics-400, the model achieves 57.5 top-1 accuracy, an improvement of 36.8%. On Something-Something-v2, the F4D model yields a 66.3 and 69.8 in top-1 accuracy with an improvement of 39.5% when pretrained on ImageNet and 43% improvement in top-1 accuracy when pretrained on ImageNet and Kinetics-400 respectively.

When the F4D model is trained on ImageNet and Kinetics-400 using 16 frames on Something-Something-v1, the F4D model achieves a 58.4 top-1 accuracy. This shows a 7.7% (50.7 vs 58.4) and 6.1% (52.3 vs 58.4) improvement in accuracy when compared with STM [34] and TEA [35] respectively. The above results show that the F4D model is capable of learning strong temporal relationships in the videos in these datasets. When the F4D model is compared to V4D using 8 frames on Something-Something-v1, the F4D model shows a 4.5% (50.4 vs 54.9) and 7.1% (50.4 vs 57.5) improvement in top-1 accuracy when pretrained on ImageNet alone and, on ImageNet and Kinetics-400 respectively. This shows that the 4D factorization and the attention modules added in the residual block of the F4D model can capture more complex inter-clip interactions and finer long-range temporal structures in the underlying video.

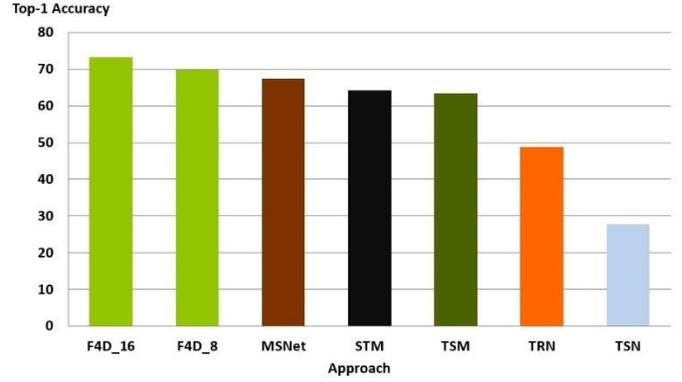

Fig. 6. Performance of F4D on Something-Something v2 compared with state-of-the-art approaches.

### D. Results on Scene-Focused Datasets

In this section, we compare the proposed F4D approach with the state-of-the-art approaches on scene-focused datasets including Kinetics-400, UCF101 and HMDB51. The videos representing most actions in these datasets are short and can be recognized by static appearance without considering temporal relationships. Furthermore, the background information contributes heavily towards deciding the action class in most of these videos.

Fig. 7 shows the results of the F4D model and other approaches on the Kinetics-400 dataset. When comparing the F4D model with STM [34] and TEA [35], F4D model shows a performance improvement of 7.5% and 5.1% respectively. Moreover, it outperforms MSNET [43] by 4.8% and V4D by 3.8%. Although the F4D model is designed specifically for temporal focused action recognition, it shows competitive results when compared to state-of-the-art methods.

Fig. 8 highlights the results on the UCF-101 and HMDB51 datasets. We follow [4] in adopting the three training/testing splits for evaluation. The F4D model was pretrained on ImageNet and Kinetics-400. In both experiments, we set $U=4$ and use 16 frames during training. Our F4D model achieves 98.2 and 84.3 accuracy on UCF101 and HMDB51 datasets respectively.

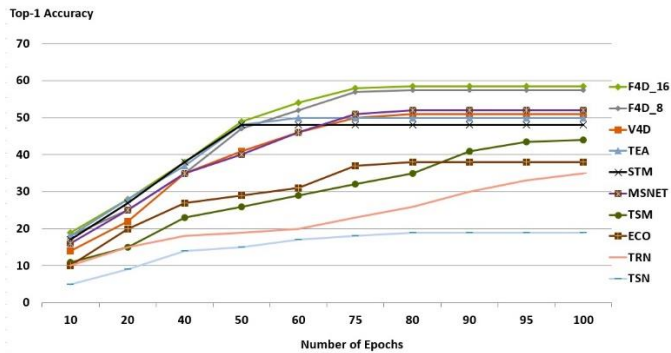

Fig. 5. Performance of F4D on Something-Something v1 compared with state-of-the-art approaches.

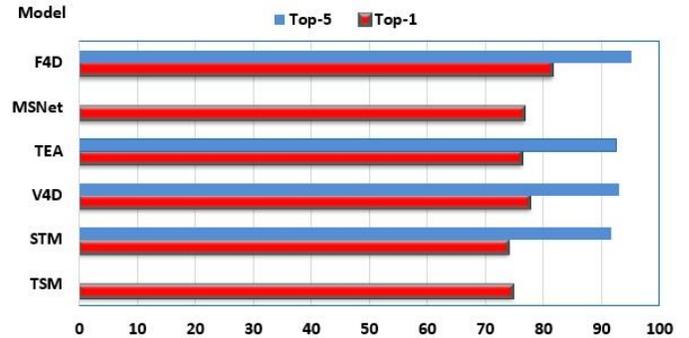

Fig. 7. Performance of the F4D model on Kinetics-400.

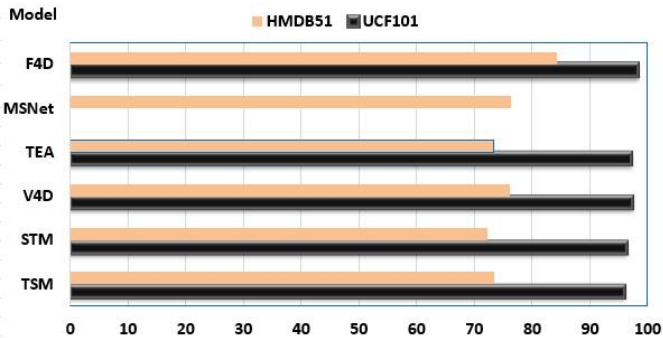

Fig. 8. Performance of the F4d model on UCF101 and HMDB51

*E. Runtime Analysis*

In this section, we compare the proposed F4D architecture with the V4D CNN. Our F4D architecture achieves better results than the V4D CNN on several benchmark datasets. Table 1 shows the model complexity and accuracy of F4D and V4D on Something-Something v1 dataset. We follow [34] to evaluate the FLOPs and speed of our architecture. We equally sample 8 or 16 frames from a video and then apply the center crop. Moreover, for speed we use a batch size of 16. All evaluations are conducted using two Tesla P100-PCIE-16GB GPU. As seen in Table 1, F4D improves the accuracy by 7.1% while achieving 2.3x less FLOPs (72G vs 167G). Moreover, our F4D gains more accuracy with 1.37x faster speed. These results demonstrate the effectiveness of the proposed factorization and attention modules in learning better and refined long-range spatiotemporal representation with less FLOPs, more speed, and a very limited increase in the number of parameters.

*F. Ablation Study*

In this section, we evaluate our F4D model on the Something-Something datasets given different scenarios. All models used in this section are pretrained on ImageNet and Kinetics-400.

**Location of F4D Blocks.** In this experiment, we study the impact of adding the F4D residual block in different positions within the F4D network. In these experiments, we fix $U=4$ and use 8 frames during training. As shown in Table 2, adding F4D blocks at conv2, conv3, conv4 or conv5 layers yields better top-1 accuracy. Adding an F4D residual block at the conv1 layer does not have a big impact which means that the short-long term features need to be refined by the earlier layers first to yield more meaningful representations. We found that adding F4D blocks from conv2 to conv5 yields the best results.

TABLE I. Model complexity of F4D compared to V4D using single crop.

| Approach | Frames | Top1 | FLOPs | Speed | # of param |
|---|---|---|---|---|---|
| V4D [5] | 8 | 50.4 | 167G | 38.1 V/s | 36.2M |
| F4D | 8 | 57.5 | 72G | 52.3 V/s | 36.8M |
| F4D | 16 | 58.4 | 143G | 27.5 V/s | 36.8M |

TABLE II. LOCATION OF F4D RESIDUAL BLOCKS

| Location | v1 top-1 accuracy | v2 top-1 accuracy |
|---|---|---|
| conv1 | 45.3 | 55.1 |
| conv2 | 49.9 | 57.3 |
| conv3 | 51.6 | 61.2 |
| conv4 | 52.8 | 63.2 |
| conv5 | 53.2 | 63.9 |
| conv2-3 | 54.2 | 64.3 |
| conv3-4 | 56.4 | 67.5 |
| conv2-5 | 57.5 | 69.8 |

TABLE III. IMPACT OF NUMBER OF ACTION UNITS FOR TRAINING

| $U_{train}$ | V1 top-1 accuracy | V2 top-1 accuracy |
|---|---|---|
| 3 | 56.8 | 69.1 |
| 4 | 57.5 | 69.8 |
| 5 | 57.9 | 70.3 |
| 6 | 58.3 | 70.5 |
| 7 | 58.5 | 70.7 |

**Number of action units $U$ used for training.** In this experiment, we observe the change in the value of $U$ during training and we found that the value of $U$ have a significant impact on overall performance. Although we anticipated obtaining higher performance figures, the videos in Something-Something datasets are relatively short and have one single and continuous action, and the action does not involve many stages. We argue that the effect of higher $U$ values will be more visible when using longer untrimmed videos during training.

**Impact of Attention Modules.** In this experiment, we study and verify the contributions of each attention module added in the proposed F4D model. We compare the results of each individual attention module and the various combinations of these attention modules. As seen in Table 4, TA+CA+STA achieves the best top-1 accuracy and outperforms the model that has no attention by 5.8% on Something-Something v1 and 9.6% on something-something v2. By combining all the attention modules, the F4D model was able to learn richer short-long term motion and spatiotemporal features.

TABLE IV. IMPACT OF ATTENTION MODULES

| Modules | v1 top-1 accuracy | v2 top-1 accuracy |
|---|---|---|
| No Attention | 51.7 | 60.2 |
| CA | 52.5 | 61.3 |
| STA | 53.8 | 63.2 |
| CA+STA | 54.2 | 65.0 |
| TA | 54.0 | 64.6 |
| TA+CA | 55.3 | 65.4 |
| TA+CA+STA | 57.5 | 69.8 |

TABLE V. COMPARISON WITH OTHER ATTENTION MODULES

| Modules | v1 top-1 accuracy | v2 top-1 accuracy |
|---|---|---|
| SE [24] | 52.1 | 60.9 |
| CBAM [39] | 52.9 | 62.1 |
| STM Block [34] | 53.9 | 64.8 |
| TEA Block [35] | 54.3 | 65.5 |
| TA+CA+STA | 57.5 | 69.8 |

**Comparison with other attention modules.** We compare the proposed TA and STA attention modules with two state-of-the-art attention modules namely SE [24] and CBAM [39]. Both attention modules can improve the performance by making the network focus on the distinctive object features by incorporating finer channel-wise attention, and the spatial module in CBAM can make the model concentrate on the spatial ROIs. First, we remove the proposed TA, CA, and STA modules in the F4D model and insert the SE module in the 3D space and compute the top-1 accuracy for both Something-Something datasets. In the second trial, we insert the CBAM instead and observe the improvement over the SE module. As illustrated in Table 5, our proposed combination of TA, CA and STA modules improves the performance significantly as both proposed attention modules exploit short term and long-term temporal relationships unlike SE and CBAM modules that do not take temporal modelling into account.

## V. CONCLUSION

In this paper, we presented an effective yet simple framework for video level representation learning namely F4D, to model both short-range motion and long-range temporal dependency at a large scale. We add the F4D residual blocks within the ResNet architecture to build the F4D pipeline. An F4D residual block performs the factorized 4D convolutional neural network which learns complex inter-clip interactions and finer temporal structures. Furthermore, it applies the two proposed attention modules to the intermediate feature maps to learn richer and refined short-long term motion and spatiotemporal features. Extensive experiments have been conducted to verify the effectiveness of F4D on five action recognition benchmark datasets, where our proposed F4D achieved state-of-the-art results.